\documentclass[runningheads]{llncs}

\usepackage[numbers]{natbib}
\usepackage[utf8]{inputenc}
\usepackage[T1]{fontenc}
\usepackage{amsmath,amssymb,amsfonts}
\usepackage{graphicx}
\usepackage{textcomp}
\usepackage{url}
\usepackage{booktabs}
\usepackage{multirow}
\usepackage{array}
\usepackage{siunitx}
\usepackage{rotating}
\usepackage{tabularx}
\usepackage{float}
\usepackage{placeins}
\usepackage{algorithm}
\usepackage{algpseudocode}
\usepackage{hyperref}
\usepackage{cleveref}
\usepackage{xcolor}
\usepackage{listings}

\def\tsc#1{\csdef{#1}{\textsc{\lowercase{#1}}\xspace}}
\tsc{WGM}
\tsc{QE}
\begin{document}
\let\WriteBookmarks\relax
\def\floatpagepagefraction{1}
\def\textpagefraction{.001}

 \title{Zero-Shot versus Fine-Tuned Transformers for Sentence-Level Media Bias Detection}

\author{
Walid fahs \inst{1} \and
Omran Berjawi  \inst{2} \and
Rida Khatoun\inst{2}}

\authorrunning{W. Fahs et al.}

\institute{Lebanese University, Beirut, Lebanon
\email{walid.fahs@ul.edu.lb}
\and
Institut Polytechnique de Paris, T\'{e}l\'{e}com Paris, Palaiseau, France\
\email{\{omran.berjawi@telecom-paris.fr, rida.khatoun@telecom-paris.fr}}
\maketitle

% Here goes the abstract
\begin{abstract}
Detecting bias in news text remains challenging because biased language is often subtle. We compare five transformer models (BERT, DistilBERT, RoBERTa, DeBERTa-v3, and BART) for sentence-level media bias classification on the expert-annotated BABE corpus, evaluating NLI-based zero-shot inference for four of them and parameter-efficient (QLoRA) fine-tuning for four. Three of the four zero-shot models perform at or near chance, and fine-tuning outperforms zero-shot inference by 22 to 28 percentage points for every architecture evaluated in both settings. Among fine-tuned models, ranking does not follow model size: RoBERTa-base performs best (macro-F1 0.85), DistilBERT matches BERT, and DeBERTa-v3-large ranks last, and the same order holds on a second corpus. Fine-tuned RoBERTa's score lies in the range reported for approaches that rely on domain-adaptive or multi-task pre-training. When tested on BASIL without adaptation, BABE-trained models lose 15 to 21 percentage points. Because the two corpora differ in domain, in annotation granularity, and in whether informational bias is labelled, this drop should not be read as a measure of domain shift alone. These results indicate that NLI-based zero-shot classification, with the single hypothesis template tested, is not a viable substitute for supervised training on this task, and that scores on BABE are a poor guide to performance under other definitions of media bias.

\end{abstract}


\section{Introduction}
\label{sec:introduction}
The advent of online news has revolutionized the way people gather information through an array of diverse sources~\cite{spinde2021automated}. However, this abundance of sources has also increased readers’ exposure to slanted, non-neutral reporting, commonly referred to as media bias~\cite{spinde2021neural}. There are various types of media bias, such as the intentional use of language, framing, and even emphasis~\cite{castillo2026automated}. Media bias can shape how readers perceive events and form opinions, which motivates the development of methods to detect it automatically.

Despite growing interest from the natural language processing (NLP) community, the automatic identification of media bias remains a challenging task. The linguistic subtlety of slanted language and the scarcity of high-quality annotated corpora collectively hinder the development of accurate and generalizable bias detectors~\cite{wessel2023introducing}. Early computational approaches relied on manually engineered lexical and syntactic features~\cite{recasens2013linguistic}, but these methods demonstrated limited scalability and poor performance on fine-grained linguistic bias. The emergence of large pre-trained transformer models such as BERT~\cite{devlin2019bert} and RoBERTa~\cite{liu2019roberta} has substantially advanced the field by enabling models to capture rich contextual representations from unlabeled text, yielding consistent performance improvements in bias detection~\cite{spinde2021neural,krieger2022domain}.

More recently, NLI-based zero-shot classification has made it possible to classify text without task-specific labeled data~\cite{yin2019benchmarking}, and it has been applied to political text~\cite{laurer2024less,banik2025bridging}. That supervised models generally outperform zero-shot ones is well established; less clear is whether zero-shot inference is usable at all for sentence-level media bias, where the target property lies in word choice and framing rather than in the propositional content that NLI models are trained to assess. To our knowledge, zero-shot and fine-tuned transformers have not been compared on the BABE benchmark under a single protocol, and it is not known whether the architectures that perform best on general benchmarks also perform best on this task. A second open challenge is generalization across datasets: many existing approaches are evaluated mainly on their training distribution, leaving cross-dataset robustness comparatively underexplored---a critical limitation for real-world deployment~\cite{wessel2023introducing}.

%More recently, zero-shot classification paradigms have opened new avenues for text classification without requiring task-specific labeled data ~\cite{yin2019benchmarking}, including political bias detection \cite{banik2025bridging}. Zero-shot approaches are particularly appealing in low-resource settings; however, to our knowledge, their performance has not been systematically compared with supervised fine-tuning on the BABE benchmark. A second important open challenge is the generalization of bias detection models across datasets and domains. Many existing approaches are evaluated mainly on their training distribution, leaving cross-domain robustness comparatively underexplored—a critical limitation for real-world deployment~\cite{wessel2023introducing}.

This paper addresses these questions by evaluating five transformer architectures on BABE~\cite{spinde2021neural} under NLI-based zero-shot inference (four models) and QLoRA fine-tuning (four models), and by testing all of them on BASIL~\cite{fan2019plain} without further adaptation. Whereas prior work on BABE aims to improve fine-tuned performance, through distant supervision, domain-adaptive or multi-task pre-training, or hyperparameter search~\cite{spinde2021neural,krieger2022domain,horych2024magpie,ghosh2025bias}, this study compares architectures under one fixed protocol, adds a zero-shot setting, and tests whether the results transfer to a second corpus.

The contributions of this paper are as follows:
\begin{enumerate}
\item We show that NLI-based zero-shot inference with a generic hypothesis template is not a usable substitute for supervision on sentence-level media bias: three of the four zero-shot models perform at or near chance on BABE, and on BASIL two of them assign nearly all sentences to a single class. On this task the gap to fine-tuning (22--28 percentage points) separates a working classifier from a non-working one, rather than a stronger classifier from a weaker one.
\item We find that, under QLoRA fine-tuning on a small expert-annotated corpus with a shared learning rate, model ranking does not follow model size or general-benchmark strength: RoBERTa-base performs best, DistilBERT matches BERT-base, and DeBERTa-v3-large ranks last. The same ranking holds on BASIL, which suggests that it is not an artefact of a single test split.
\item We show that BABE-trained detectors lose 15--21 percentage points on BASIL. Because the two corpora differ in domain, in annotation granularity, and in whether informational bias is labelled, this drop should not be read as a measure of domain shift alone; rather, it indicates that BABE scores are a poor guide to how well a model detects media bias as defined elsewhere.
\item We find that direct parameter-efficient fine-tuning of RoBERTa-base, updating only low-rank adapters, reaches a macro-F1 of 0.85, within the range reported for approaches that rely on domain-adaptive or multi-task pre-training; differences in evaluation protocol make this comparison indicative only.
\end{enumerate}

The remainder of this paper is organized as follows. Section~\ref{sec:related} reviews related work. Section~\ref{sec:methodology} details the methodology. Section~\ref{sec:experiments} describes the experimental setup. Section~\ref{sec:results} presents the results. Sections~\ref{sec:discussion} and~\ref{sec:Limitations} discuss the findings and the limitations, respectively. Section~\ref{sec:conclusion} concludes the paper. 

\section{Related Work}
\label{sec:related}
The formation of high-quality annotated corpora is critical for advancing automatic media bias detection. Earlier media bias annotation corpora suffered from low inter-annotator agreement and narrow topic coverage~\cite{lim2020annotating}. For example, Lim et al.~\cite{lim2020annotating} used crowdsourcing to annotate 966 sentences with bias labels, but the Fleiss' Kappa coefficient was approximately zero. Similarly, Färber et al.~\cite{farber2020multidimensional} compiled a multi-dimensional crowdsourced dataset of over 2,000 sentences drawn from coverage of a single political event domain, and likewise reported low inter-annotator agreement. To overcome these limitations, Spinde et al.~\cite{spinde2021neural} proposed BABE, the highest-agreement expert-annotated corpus for sentence-level media bias at the time of its release (Krippendorff's $\alpha$ = 0.40), substantially exceeding the agreement of earlier datasets. MBIC ~\cite{spinde2021mbic} developed by the same group, is a precursor of BABE. More recently, Wessel et al.~\cite{wessel2023introducing} proposed MBIB, a comprehensive benchmark of 22 media bias datasets covering nine types of biases, showing that there is no single algorithm capable of handling all kinds of media bias. Finally, Raza et al.~\cite{raza2026beads} introduced BEADs, a multi-task bias dataset of approximately 50,000 gold-labeled records filtered from an aggregated corpus of roughly 3.7 million texts.

\subsection{Transformer-Based Bias Detection} 
The application of pre-trained transformer models to media bias detection has yielded substantial performance gains over feature-based approaches. Spinde et al.~\cite{spinde2021neural} demonstrated that BERT and RoBERTa, pre-trained on distantly labeled news headlines from politically partisan and neutral outlets, achieved  macro-F1 scores of 0.804 and 0.799, respectively, on BABE, far exceeding feature-based baselines. Their distant supervision approach leveraged the AllSides media bias chart to assign weak labels to large quantities of news text automatically. Building on this foundation, Krieger et al.~\cite{krieger2022domain} proposed a domain-adaptive pre-training (DAPT) approach in which RoBERTa was first intermediate-trained on the Wiki Neutrality Corpus (WNC)—a collection of 180,000 Wikipedia sentence pairs labeled for subjective bias—before fine-tuning on BABE. Their model, DA-RoBERTa, achieved an F1 score of 0.814, establishing a new state-of-the-art at the time. 	Ghosh et al. ~\cite{ghosh2025bias} subsequently showed that fine-tuning a standard RoBERTa-base model directly on BABE, without domain-adaptive pre-training but with careful hyperparameter tuning and early stopping,  achieved a  macro-F1 of 0.926 under stratified 5-fold cross-validation and outperformed DA-RoBERTa (0.836 under the same protocol) with statistical significance. Their attention-based analysis revealed that domain-adaptive overfitting on Wikipedia-style text caused DA-RoBERTa to over-attend to politically charged tokens while missing framing-based cues.

\subsection{Multi-Task and Large-Scale Pre-Training Approaches} 
Recognizing that media bias is a composite, multi-faceted construct, Horych et al.~\cite{horych2024magpie} introduced MAGPIE, a large-scale multi-task learning approach that pre-fine-tuned a RoBERTa encoder on 59 bias-related tasks spanning 1,210,084 labeled sentences organized into nine task families, compiled into the Large Bias Mixture (LBM) collection. MAGPIE achieved an F1 score of 0.841 on BABE, a relative improvement of 3.3\% over the previous state-of-the-art, and required only 15\% of the training steps needed by single-task approaches. Their task-family transfer analysis showed that sentiment analysis and emotionality tasks provided the most consistent positive transfer to bias detection, while scaling the number of tasks yielded the best overall results.

\subsection{Zero-Shot Classification for Bias Detection} 
Zero-shot classification has emerged as a promising paradigm for NLP tasks where labeled data is scarce or unavailable~\cite{yin2019benchmarking}. In this paradigm, models leverage knowledge acquired during pre-training or natural language inference (NLI) fine-tuning to classify inputs without task-specific supervision. Banik et al.~\cite{banik2025bridging} conducted a comparative study of zero-shot and few-shot prompting across BERT, RoBERTa, and GPT-series models on the FIGNEWS dataset for political bias detection, finding that among encoder-only models, RoBERTa achieved the strongest supervised performance, while generative models showed stronger zero-shot alignment with human annotations. The BEADs benchmark~\cite{raza2026beads} also compared zero-shot and fine-tuned large language models, demonstrating that encoder-only models consistently outperformed autoregressive models under fine-tuning, while large language models showed inconsistent zero-shot behavior across demographic groups. Yin et al.~\cite{yin2019benchmarking} introduced a general-purpose NLI-based zero-shot text classification framework, demonstrating that NLI-fine-tuned models can be repurposed as classifiers by reformulating labels as natural language hypotheses—a paradigm directly applicable to bias detection. 
Similarly, Laurer et al.~\cite{laurer2024less} evaluated NLI-based classifiers across eight political text classification tasks, showing that a BERT-NLI model fine-tuned on only 100 to 2,500 examples outperforms classical supervised models by 10.7 to 18.3 percentage points on average, and that its prior task knowledge is particularly valuable under data scarcity and class imbalance. Despite these advances, a comparison of zero-shot and fine-tuned transformers on the BABE benchmark under a single, consistent protocol has not, to our knowledge, been reported.

\subsection{Generalization in Bias Detection} 
Cross-domain generalizability has long been recognized as an important problem in media bias detection research. As Wessel et al.~\cite{wessel2023introducing} note, models designed to detect one kind of bias, or even one dataset, rarely generalize to other kinds/datasets, suggesting no unified framework for detecting media bias. Fan et al.~\cite{fan2019plain} presented BASIL, a dataset of 300 news articles annotated at the span level for lexical and informational bias.

\section{Methodology} 
\label{sec:methodology}
This section presents the methodology for sentence-level media bias detection. It is organized around three core components: task formulation, which defines the classification objective and its formal basis; model description, which characterizes the five transformer architectures selected for evaluation and justifies their inclusion; and classification paradigms, which articulate the two adaptation strategies—zero-shot inference and supervised fine-tuning—applied to the selected architectures.

\subsection{Task Definition} 
The task of detecting media bias at a sentence level is formulated as a binary classification problem. For a sentence set $\mathcal{X}$ containing sentences $x_i \in \mathcal{X}, i = 1, \ldots, N$, the aim is to label them using $y_i \in \{0, 1\}$, wherein $y_i=1$ means that the sentence is biased, while $y_i = 0$ implies that the sentence is unbiased. This formulation follows ~\cite{spinde2021neural} and ~\cite{ghosh2025bias}. The loss function used for training is:

\begin{equation}
L = - \frac{1}{N} \sum_{i=1}^{N} \sum_{k \in \{0,1\}} y_{i,k} \log(\hat{y}_{i,k})
\label{eq:loss}
\end{equation}

where $y_{i,k}$ is the ground-truth indicator (1 if sample i belongs to class k, 0 otherwise), and $\hat{y}_{i,k}$  is the predicted probability for class k. 

Since the label space contains two classes, Equation~\eqref{eq:loss} is the two-class form of categorical cross-entropy and is equivalent to the standard binary cross-entropy objective.

Sentence-level detection is the standard setting in recent media bias work~\cite{spinde2021neural,krieger2022domain,ghosh2025bias}, as the sentence is a common unit of analysis that can be aggregated to paragraph or article level.

\subsection{Transformer-Based Models} 
Building upon the prior work of Spinde et al.~\cite{spinde2021neural}, Krieger et al.~\cite{krieger2022domain}, and Ghosh et al.~\cite{ghosh2025bias}, five transformer architectures are evaluated, spanning the encoder and encoder-decoder families used in prior bias-detection work.

BERT~\cite{devlin2019bert}, DistilBERT~\cite{sanh2019distilbert}, and RoBERTa~\cite{liu2019roberta} constitute the encoder-only backbone of this evaluation. BERT introduces fully bidirectional contextual representations through masked language modeling (MLM) and next sentence prediction (NSP) pre-training, providing strong sentence-level representations via the [CLS] token. DistilBERT distills BERT into a 40\% smaller, 60\% faster model while retaining approximately 97\% of its performance~\cite{sanh2019distilbert}, offering a computationally efficient classification baseline. For zero-shot inference, the distilbert-base-uncased-mnli checkpoint is used. RoBERTa refines BERT by removing NSP, adopting dynamic masking, and training on a substantially larger corpus including news and web text—modifications that yield stronger contextual representations and have produced strong results in prior media bias work~\cite{krieger2022domain,ghosh2025bias,horych2024magpie}.  For zero-shot inference, the roberta-large-mnli checkpoint is used, an NLI-fine-tuned variant that enables entailment-based classification without task-specific training. 

DeBERTa~\cite{he2020deberta} and BART~\cite{lewis2020bart} extend the evaluation to architectures with distinct inductive biases. DeBERTa introduces disentangled attention, representing each token with separate content and position vectors and computing attention from content-to-content, content-to-position, and position-to-content interactions. For zero-shot inference, the deberta-v3-large-finetuned-mnli checkpoint is used. BART~\cite{lewis2020bart}, combines a bidirectional encoder with an autoregressive decoder trained via a denoising objective. It is evaluated only in the zero-shot setting, using the bart-large-mnli checkpoint, the standard model for NLI-based zero-shot classification. Fine-tuning BART was left out to keep the fine-tuned comparison focused on encoder-only models and within our computational budget.

\subsection{Classification Approaches: Zero-Shot and Fine-Tuning}
\label{subsec:approaches}
Zero-shot classification is a paradigm in which a pre-trained model performs the target task without any task-specific parameter updates, relying entirely on knowledge acquired during pre-training and NLI fine-tuning~\cite{yin2019benchmarking}. This approach is operationalized through the NLI entailment framework: each sentence is treated as a premise, and each candidate label is inserted into a hypothesis template (see Section~\ref{subsec:impl}). The label whose hypothesis receives the highest entailment score is assigned as the prediction. This setting requires no labeled data and estimates each model’s ability to detect bias without supervision, under the chosen hypothesis template. All NLI-fine-tuned checkpoints (roberta-large-mnli, deberta-v3-large-finetuned-mnli, bart-large-mnli, and distilbert-base-uncased-mnli) are applied directly to the BABE test split and to BASIL without any further adaptation.

Supervised fine-tuning adapts a pre-trained transformer to the bias classification task by updating model parameters on labeled training examples using the cross-entropy objective defined in Equation~\eqref{eq:loss}. Unlike zero-shot inference, fine-tuning allows the model to acquire task-specific representations by directly optimizing on the distribution of the target dataset. During fine-tuning, the default sequence-classification head of each checkpoint is applied to the representation of the first token. Under the QLoRA configuration described in Section~\ref{subsec:impl}, the pre-trained weights are quantized to 4 bits and kept frozen; only the low-rank adapter matrices and the classification head are updated. Model selection is performed based on the best validation  macro-F1 score observed across training epochs, ensuring that the final evaluation uses the checkpoint with the best validation macro-F1 rather than an arbitrary stopping point.

Table~\ref{tab:configs} lists the checkpoints used in each setting. Zero-shot checkpoints are additionally fine-tuned on MNLI, which the NLI approach requires. The DistilBERT and DeBERTa-v3 pairs are size-matched, whereas RoBERTa uses a larger checkpoint for zero-shot inference; BERT and BART are evaluated in one setting only. The paired results therefore compare specific model configurations rather than isolating the effect of the learning approach, with the size-matched pairs giving the cleaner comparison.

\begin{table}[htbp]
\centering
\caption{Checkpoints used in each setting. All zero-shot checkpoints are fine-tuned on MNLI.}
\label{tab:configs}
\small
\begin{tabular}{lll}
\toprule
\textbf{Model} & \textbf{Zero-shot checkpoint} & \textbf{Fine-tuned checkpoint} \\
\midrule
BERT       & --                              & bert-base-uncased \\
DistilBERT & distilbert-base-uncased-mnli    & distilbert-base-uncased \\
RoBERTa    & roberta-large-mnli              & roberta-base \\
DeBERTa-v3 & deberta-v3-large-finetuned-mnli & deberta-v3-large \\
BART       & bart-large-mnli                 & -- \\
\bottomrule
\end{tabular}
\end{table}
 
\section{Experimental Setup}
\label{sec:experiments}
 
\subsection{Datasets}

\subsubsection{BABE} 
This dataset~\cite{spinde2021neural} is used as the main data resource for training and testing for in-domain tasks. It includes 3,700 sentences gathered from 14 U.S. media sources concerning 12 controversial topics. All the sentences have been annotated at the word level and sentence level by domain experts to produce binary labels (biased/non-biased) on the sentence level, as well as lists of words contributing to bias on an instance level. There is also structured metadata provided for each instance: the source, topic category, news article, and the label of political sentiment, where available.

\subsubsection{BASIL} 
BASIL~\cite{fan2019plain} is used as the external out-of-distribution benchmarking dataset for generalization testing. This dataset contains 300 news articles that have been classified into 100 event-consistent trios. Each trio includes news reports on the same political event, but from three different American media organizations that espouse varying ideological stances:  the New York Times, Fox News, and the Huffington Post. Articles are annotated at the span level for two complementary bias types: lexical bias, arising from evaluative word choice, and informational bias, arising from the selective presentation of facts. In total, BASIL contains 1,727 annotated bias spans across all articles. For this study, span-level annotations are converted to binary sentence-level labels: a sentence is labeled as biased if it contains at least one annotated bias span, and as non-biased otherwise. This conversion aligns BASIL with BABE’s sentence-level label space, so that models trained on BABE can be evaluated on BASIL without modification. It is, however, an approximation: the two corpora define bias differently, and sentence-level labels discard span information (see Sections~\ref{subsec:disc-generalization} and~\ref{sec:Limitations}).

\subsection{Implementation Details}
\label{subsec:impl}
All models are implemented using the Hugging Face Transformers library, with model weights instantiated from their official pre-trained checkpoints hosted on the Hugging Face Model Hub. The experimental framework is built on Python 3.10 and PyTorch 2.1.0. All experiments are executed on a Kaggle Notebook environment equipped with an NVIDIA Tesla T4 GPU (16 GB VRAM). All default tokenizers associated with each pre-trained checkpoint are used, with a maximum sequence length of 128 tokens.

\subsubsection{Zero-shot}
NLI-fine-tuned checkpoints (roberta-large-mnli, deberta-v3-large-finetuned-mnli, bart-large-mnli, and distilbert-base-uncased-mnli) are loaded from the Hugging Face Model Hub and applied without any parameter updates. Inference uses the Hugging Face zero-shot-classification pipeline with the candidate labels “biased” and “not biased” and the pipeline's default hypothesis template, ``This example is \{\}''. In single-label mode, the entailment logits of the two hypotheses are normalized with a softmax, and the label with the higher score is assigned as the prediction. The same procedure is applied to the BABE test split and to the converted BASIL sentences.

\subsubsection{Fine-tuning} 
The roberta-base, bert-base-uncased, distilbert-base-uncased, and deberta-v3-large checkpoints are fine-tuned using the following QLoRA \cite{dettmers2023qlora} configuration: 4-bit quantization, $r = 32$, $\alpha = 64$, and dropout = 0.1. QLoRA is employed to enable memory-efficient fine-tuning and to provide a consistent fine-tuning framework across all architectures.  All models are optimized with AdamW (learning rate = $2 \times 10^{-5}$, batch size = 32). An 80/20 stratified train-test split is applied to BABE to preserve label distribution, with 10\% of the training partition held out as a validation set. Models are trained for a maximum of 10 epochs using early stopping with a patience of 3 epochs based on validation  macro-F1. The best checkpoint is frozen and evaluated exactly once on the held-out test split.

\subsection{State-of-the-Art Comparison}
To assess the performance of this approach relative to prior work, the best-performing model from the zero-shot and fine-tuning evaluation is compared against four previously published systems on the BABE benchmark: the distant supervision approach of Spinde et al. ~\cite{spinde2021neural}, the domain-adaptive pre-trained DA-RoBERTa of Krieger et al.~\cite{krieger2022domain}, the large-scale multi-task model MAGPIE of Horych et al.~\cite{horych2024magpie}, and the fine-tuned bias-detector of Ghosh et al.~\cite{ghosh2025bias}.

\subsection{Evaluation Metrics}
 macro-F1 score is reported as the primary evaluation metric for both the in-domain (BABE) and out-of-distribution (BASIL) settings.  macro-F1 is computed as the unweighted average of per-class F1 scores, treating biased and non-biased classes equally.

\section{Results}
\label{sec:results}
\subsection{In-Domain Performance on BABE}
Table~\ref{tab:babe} summarizes the macro-F1 scores of all models on BABE. Among fine-tuned models, RoBERTa achieves the highest macro-F1 (0.85), followed by BERT and DistilBERT (0.78 each) and DeBERTa-v3 (0.75). Model size does not predict this ranking: DistilBERT, which is 40\% smaller, matches BERT, and the largest model, DeBERTa-v3-large, ranks last; possible reasons are discussed in Section~\ref{sec:discussion}. Under zero-shot inference, only RoBERTa (0.63) performs clearly above chance, whereas BART, DeBERTa-v3 and DistilBERT score 0.49--0.50. DeBERTa-v3-large is among those at chance although it outperforms RoBERTa-large on MNLI itself~\cite{he2021debertav3}, so stronger NLI performance does not translate into better zero-shot bias detection. The gap between fine-tuned and zero-shot performance is 28 percentage points for DistilBERT and 25 for DeBERTa-v3, both size-matched pairs, and 22 for RoBERTa, where the zero-shot model is the larger of the two (Section~\ref{subsec:approaches}).

\begin{table}[htbp]
\centering
\caption{In-domain performance on BABE. FT = fine-tuned; ZS = zero-shot.}
\label{tab:babe}
\begin{tabular}{lc}
\toprule
\textbf{Model} & \textbf{ macro-F1} \\
\midrule
DistilBERT (FT) & 0.78 \\
DistilBERT (ZS) & 0.50 \\
RoBERTa (FT)    & \textbf{0.85} \\
RoBERTa (ZS)    & 0.63 \\
DeBERTa (FT)    & 0.75 \\
DeBERTa (ZS)    & 0.50 \\
BART (ZS)       & 0.49 \\
BERT (FT)       & 0.78 \\
\bottomrule
\end{tabular}
\end{table}

\subsection{State-of-the-Art Comparison on BABE}
Table~\ref{tab:sota} places fine-tuned RoBERTa alongside four previously published systems evaluated on BABE. These figures are not directly comparable: the studies differ in how BABE is partitioned (a single stratified split here, versus 5-fold cross-validation in Ghosh et al.~\cite{ghosh2025bias}), in random seeds, and in how the model is adapted. The size of such effects is illustrated by DA-RoBERTa, which scores 0.814 as originally reported~\cite{krieger2022domain} and 0.836 when re-evaluated by Ghosh et al. under their protocol. Differences of a few points between rows should therefore not be read as a ranking.

With this caveat, fine-tuned RoBERTa (0.85) falls within the range reported for approaches based on distant supervision, domain-adaptive pre-training or multi-task learning (0.80--0.84), while using only parameter-efficient fine-tuning on BABE. It remains below the 0.93 of Ghosh et al., who also fine-tune RoBERTa-base directly but update all parameters and tune hyperparameters; whether this gap reflects these choices or the evaluation protocol cannot be determined from the reported results.

\begin{table}[htbp]
\centering
\caption{Reported results on BABE (protocols differ). FT = fine-tuning.}
\label{tab:sota}
\small
\begin{tabular}{llc}
\toprule
\textbf{System} & \textbf{Adaptation} & \textbf{Macro-F1} \\
\midrule
Spinde et al.~\cite{spinde2021neural}   & Distant supervision + full FT     & 0.80 \\
Krieger et al.~\cite{krieger2022domain} & DAPT on WNC + full FT             & 0.81 \\
Horych et al.~\cite{horych2024magpie}   & Multi-task pre-training + full FT & 0.84 \\
Ghosh et al.~\cite{ghosh2025bias}       & Full FT + hyperparameter search   & 0.93 \\
\midrule
This work (RoBERTa)                      & QLoRA FT, no hyperparameter search & 0.85 \\
\bottomrule
\end{tabular}
\end{table}
 
\subsection{Cross-Dataset Generalization to BASIL}
Table~\ref{tab:basil} reports results on BASIL for the fine-tuned models trained on BABE and for the zero-shot models, in both cases without further adaptation. All fine-tuned models lose 15 to 21 percentage points relative to BABE, but their ranking is unchanged: RoBERTa (0.70) remains best, followed by DistilBERT (0.62), BERT (0.61) and DeBERTa-v3 (0.54), which shows the largest drop. Zero-shot results diverge more: RoBERTa loses 9 points and DeBERTa-v3, already near chance, 3, whereas DistilBERT (0.32) and BART (0.24) fall below the level expected from random guessing, a pattern consistent with assigning most sentences to a single class. Because fine-tuned RoBERTa drops more than its zero-shot counterpart, the advantage of fine-tuning narrows from 22 to 16 points. Zero-shot RoBERTa ties fine-tuned DeBERTa-v3 at 0.54; otherwise every fine-tuned model scores above every zero-shot model, and for each architecture evaluated in both settings the fine-tuned variant outperforms its zero-shot counterpart on both corpora. Possible causes of the drop are discussed in Section~\ref{subsec:disc-generalization}.
 
\begin{table}[htbp]
\centering
\caption{Cross-dataset evaluation on BASIL. FT = fine-tuned; ZS = zero-shot.}

\label{tab:basil}
\begin{tabular}{lc}
\toprule
\textbf{Model} & \textbf{ macro-F1} \\
\midrule
DistilBERT (FT) & 0.62 \\
DistilBERT (ZS) & 0.32 \\
RoBERTa (FT)    & \textbf{0.70} \\
RoBERTa (ZS)    & 0.54 \\
DeBERTa (FT)    & 0.54 \\
DeBERTa (ZS)    & 0.47 \\
BART (ZS)       & 0.24 \\
BERT (FT)       & 0.61 \\
\bottomrule
\end{tabular}
\end{table}

\section{Discussion}
\label{sec:discussion}

\subsection{Zero-shot NLI inference does not work for this task}
Three of the four zero-shot models perform at or near chance on BABE, and the gap to fine-tuning is at least as large in the size-matched pairs as in the RoBERTa pair. One possible explanation, which we have not tested, is that NLI models are trained to judge whether a hypothesis follows from what a premise asserts, whereas the bias annotated in BABE lies largely in word choice that leaves the assertion unchanged; a hypothesis such as ``This example is biased'' is then not something the premise entails. This is consistent with the observation that DeBERTa-v3-large, the strongest of the four checkpoints on MNLI, performs at chance. Why roberta-large-mnli alone rises above chance is not explained by our results. In practice, generic NLI zero-shot classification is not a usable fallback when labels are unavailable; consistent with prior work~\cite{yin2019benchmarking,laurer2024less}, even a small labeled set is likely the better investment.

\subsection{Model size does not predict performance}
Under QLoRA fine-tuning, RoBERTa-base performs best, DistilBERT matches BERT-base despite being 40\% smaller, and DeBERTa-v3-large ranks last; the same ranking holds on BASIL. Two explanations are plausible but untested. RoBERTa's pre-training corpus includes a large share of news text, which may match the register of BABE. DeBERTa-v3-large, the largest model, may be under-trained with a single learning rate and about 2,700 training sentences under low-rank adaptation. With small expert-annotated data and parameter-efficient fine-tuning, a base-size model is therefore a reasonable default, and DistilBERT a low-cost alternative with no loss in our setting. Fine-tuned RoBERTa also reaches the range reported for domain-adaptive and multi-task pre-training~\cite{krieger2022domain,horych2024magpie}, consistent with Ghosh et al.~\cite{ghosh2025bias}; because these results come from different protocols, this comparison is indicative only.

\subsection{Performance on BABE does not transfer to BASIL}
\label{subsec:disc-generalization}
On BASIL, every fine-tuned model scores 15 to 21 percentage points lower than on BABE, while their ranking is unchanged; the advantage of fine-tuned over zero-shot RoBERTa narrows from 22 to 16 points. The cause of the drop cannot be established from the present design, because the two corpora differ in three ways at once: they cover different articles, events and topics; BASIL also annotates informational bias, which arises from the selective presentation of facts rather than from word choice, whereas BABE targets lexical bias; and BASIL's span-level annotations were converted to sentence-level labels, which may add label noise. Their relative contributions are unknown and need not be additive. What the results do show is that a score on BABE is a poor guide to how well a model detects bias as another corpus defines it, so BABE results should not be read as evidence of general media bias detection.

\section{Limitations and Future Directions}
\label{sec:Limitations}

\textbf{Experimental design.} All results come from a single stratified split and a single random seed, so differences of one or two percentage points should not be over-interpreted. Fine-tuning used one learning rate ($2\times10^{-5}$), lower than is typical for low-rank adapters, and no hyperparameter search, which may underestimate some models, particularly DeBERTa-v3-large. The zero-shot and fine-tuned configurations are also not fully controlled (Section~\ref{subsec:approaches}). Future work will use repeated runs, a hyperparameter search, and size-matched checkpoint pairs for every architecture, including BART.

\textbf{Zero-shot setting.} Zero-shot inference used a single hypothesis template, and instruction-tuned large language models were not evaluated; both are needed before drawing conclusions about zero-shot approaches beyond NLI.

\textbf{Cross-dataset evaluation.} Converting BASIL's span-level annotations to sentence-level labels discards span information, and the design cannot separate domain shift from differences in bias definition. Evaluating on lexical-bias spans only, and training across several corpora and bias types, for example through multi-task learning~\cite{horych2024magpie} or adversarial domain adaptation, are natural next steps.

\textbf{Comparison with prior work.} The comparison in Table~\ref{tab:sota} combines results obtained under different splits and protocols; a controlled re-evaluation of all baselines remains future work.

\textbf{Scope.} The sentence-level formulation cannot capture discourse-level phenomena such as framing and omission, which are better studied at the paragraph or article level. Finally, no qualitative error analysis was performed, in part because BABE lacks bias-type labels that would allow errors to be grouped by category; such an analysis would show which forms of bias each model misses.

\section{Conclusion}
\label{sec:conclusion}
This paper compared five transformer models (four under NLI-based zero-shot inference and four under QLoRA fine-tuning) for sentence-level media bias detection. Fine-tuned RoBERTa achieved the best performance on both test sets, reaching a macro-F1 of 0.85 on BABE without domain-adaptive pre-training or multi-task learning. Zero-shot inference with the hypothesis template tested reached at most 0.63, indicating that, for the configurations evaluated, task-specific supervision remains essential. Transferring from BABE to BASIL reduced macro-F1 by 15 percentage points for the best model; because the corpora differ in domain, bias definition and annotation granularity, the sources of this drop cannot be separated in the present design. Future work will evaluate instruction-tuned large language models in zero- and few-shot settings, use repeated runs to quantify variance, and separate lexical from informational bias in cross-dataset evaluation.

\bibliographystyle{splncs04}
\bibliography{references}

\end{document}